%% file: main.tex
\documentclass[runningheads]{llncs}

\usepackage{eccv}

\usepackage{eccvabbrv}

\usepackage{graphicx}
\usepackage{booktabs}
\usepackage{algorithm}
\usepackage{algpseudocode}
\usepackage[accsupp]{axessibility}  

\usepackage{hyperref}

\usepackage{orcidlink}

\input{macros}
\begin{document}

\title{V-Trans4Style: Visual Transition Recommendation for Video Production Style Adaptation} 

\titlerunning{V-Trans4Style}

\author{Pooja Guhan\inst{1,2}\orcidlink{0000-0003-1551-8163} \and
Tsung-Wei Huang\inst{2}\orcidlink{0000-0002-1478-2678} \and
Guan-Ming Su\inst{2}\orcidlink{0000−0002−3118−5904}\and \\ Subhadra Gopalakrishnan\inst{2}\orcidlink{0009-0007-7878-6858}\and
Dinesh Manocha\inst{1}\orcidlink{0000-0001-7047-9801}}

\authorrunning{Guhan, Pooja et al.}

\institute{University of Maryland, College Park MD 20740, USA \and
Dolby Laboratories, Sunnyvale CA 94085, USA}

\maketitle
\begin{abstract}
  We introduce V-Trans4Style, an innovative algorithm tailored for dynamic video content editing needs. It is designed to adapt videos to different production styles like documentaries, dramas, feature films, or a specific YouTube channel's video-making technique. Our algorithm recommends optimal visual transitions to help achieve this flexibility using a more bottom-up approach. We first employ a transformer-based encoder-decoder network to learn recommending temporally consistent and visually seamless sequences of visual transitions using only the input videos. We then introduce a style conditioning module that leverages this model to iteratively adjust the visual transitions obtained from the decoder through activation maximization. We demonstrate the efficacy of our method through experiments conducted on our newly introduced AutoTransition++ dataset. It is a 6k video version of AutoTransition Dataset that additionally categorizes its videos into different production style categories. Our encoder-decoder model outperforms the state-of-the-art transition recommendation method, achieving improvements of 10\% to 80\% in Recall@K and mean rank values over baseline. Our style conditioning module results in visual transitions that improve the capture of the desired video production style characteristics by  an average of around 12\% in comparison to other methods when measured with similarity metrics. We hope that our work serves as a foundation for exploring and understanding video production styles further. \href{https://gamma.umd.edu/v-trans4style}{[Project Website]}
  \keywords{video transitions \and video style \and transition recommendation \and video editing}
\end{abstract}

\section{Introduction}
\label{sec:intro}

With the growing popularity and consumption of digital content, creating high-quality videos still remains a time-consuming, expensive, and highly creative task. To meet the ever-growing demand for diverse and captivating multimedia experiences, the strategic act of \textit{adapting} videos into various production styles emerges as an invaluable resource. \textit{Video production styles} encompass distinct approaches and techniques employed in crafting videos. These styles aim to convey specific messages, elicit emotions, or achieve particular aesthetics. While there is no finite set of video styles, they can be attributed to genre-specific approaches, such as those used in documentaries, feature films, news/reportage, drama, and vlogs. Additionally, video styles can be specific to famous directors and content creators, or they may reflect an organization's distinctive marketing styles~\cite{alghamdi2024makes}.
These video styles can be achieved with a mix of multiple visual elements, including camera angles, scene organization, and visual transitions~\cite{edittransformgenre}. Adapting videos into these varied styles can help content creators create videos for diverse audience preferences and consistently deliver captivating content that stands out in the competitive landscape of digital media. However, editing videos to specific styles remains a laborious task. In this work, we limit our scope of understanding the dependence of video styles on visual transitions.
\begin{figure}
    \centering
    \includegraphics[width=0.7\textwidth]{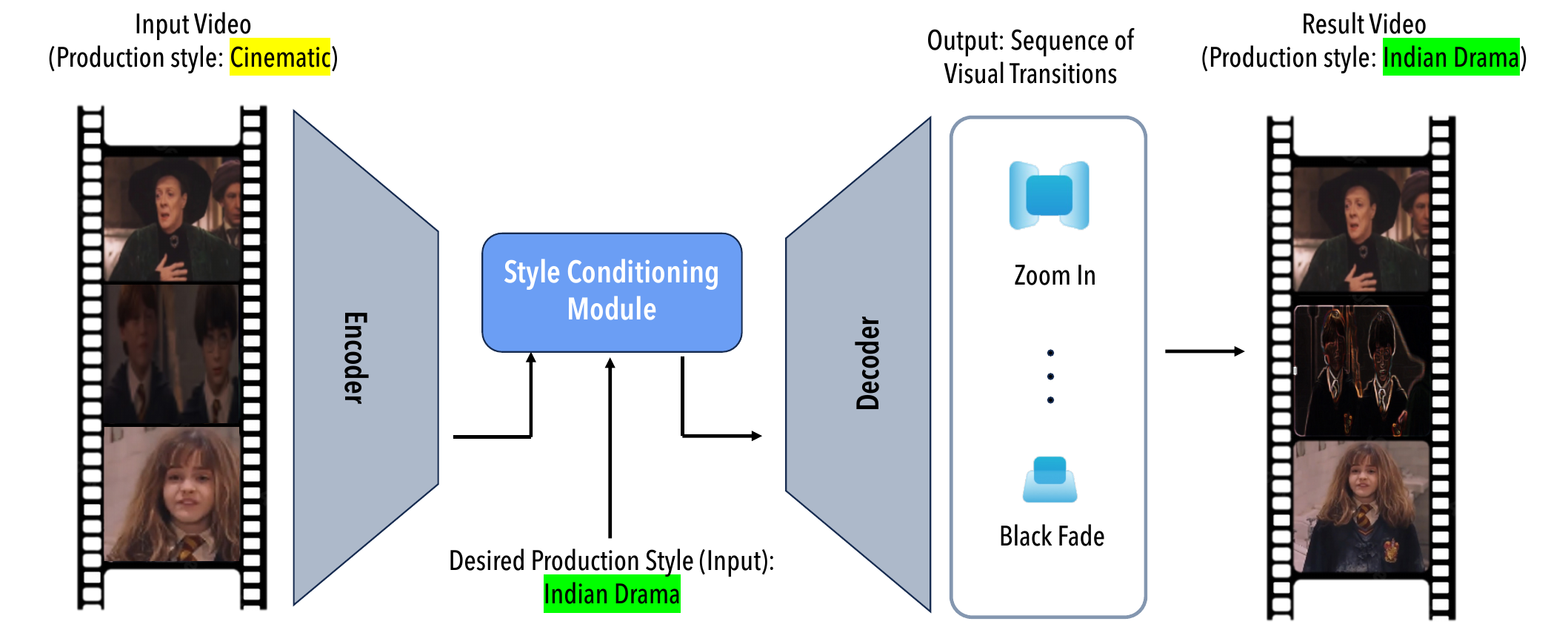}
    \caption{Our goal is to recommend the optimal visual transition sequence for enabling the adaptation of a given video to any desired production style. We propose~\modelname, a novel bottom-up approach consisting of an encoder-decoder architecture and a style conditioning module. }
    \label{fig:enter-label}
\end{figure}
\textit{Visual transitions} serve as the creative glue during post-production, seamlessly connecting video clips and enhancing the video's flow and style. These effects enable smooth shifts between shots, often introducing a visual and sometimes thematic link between them~\cite{Adobe}.
Fades, dissolves, wipes, cross-fades, etc are among the various types of transitions, and they have a significant impact on the overall look and feel of the video~\cite{zhou2023analysis}.
Today, state-of-the-art video editing softwares~\cite{cloud2014adobe}~offer capabilities for implementing these transitions in videos. For non-professionals lacking expertise in cinematography and design, the challenge lies in selecting transitions from numerous possibilities and exploring various combinations to adapt videos for different production styles. Even seasoned editors grapple with its time-consuming nature. Video templates, i.e., pre-designed and pre-structured video layouts,  available today are limited, rigid, and are not tailored for the user's video content on which it is being applied. This thereby, reduces the overall quality of the adapted video. Therefore, \textit{we introduce an automated method of recommending transitions for adapting videos for different video production styles}. 
To attain our objective, we must navigate through certain challenges. Firstly, \textit{no} publicly available dataset provides the different sets of transitions that need to be applied to the same video to adapt it for different styles. Developing such a dataset is also an extremely tedious task. Additionally, unlike measurable aspects of video production such as frame rate or resolution, video style is abstract and lacks concise formulation. Apart from the desired video style, the sequence of transitions being predicted for adapting a video should also match the dynamics and rhythm of the video content as much as possible to allow for video continuity.  

\noindent\textbf{Main Contributions:}
To overcome these challenges, we present~\modelname. Targeting video editing applications, our goal is to provide content creators with a user-friendly approach that can enable them to adapt their videos to different production styles, ultimately enhancing engagement and broadening audience appeal. Our main contributions can be summarized as follows:
\begin{enumerate}
    \item We present the first algorithm that facilitates video adaptation to various styles during video editing. The input is a video comprised of an ordered set of video clips and the name of the desired video production style. The output is the recommended sequence of visual transition classes that the user should use to adapt the input video to the desired video production style.
    \item We present a novel method that follows a bottom-up approach to recommend transitions suitable for video production style adaptation. We first train a transformer encoder-decoder architecture to learn to recommend temporally and visually consistent visual transitions. We then use our style conditioning module to perform a controlled update of its latent embedding during inference time via activation maximization. This leads to the introduction of video production style-based characteristics in the sequence of transitions obtained from the decoder. 
    \item We create~\dataname, a mini representative dataset of AutoTransition~\cite{shen2022autotransition} consisting of $6k$ videos. Additionally, in a bid to fortify our research foundation, we also include additional annotations to categorize the videos within it into five distinct video production styles. 
\end{enumerate}
We show quantitative evaluations on~\dataname. Our encoder-decoder model shows an improvement of at least $10$\%  in Recall@K and mean rank metrics as compared to the baseline. Additionally, we show that our style conditioning module can help better capture the video production style characteristics by $12$\% as compared to baselines.  

\section{Related Works}
\label{sec:related_works}
\noindent\textbf{Video Editing.}
The recent surge in the popularity of video editing is largely attributed to advancements in generative AI technologies. Within this field, four main categories of work have emerged.
The first category focuses on content manipulation, where existing visual elements are modified on a frame-by-frame basis. This empowers users to substitute specific objects with their preferred choices~\cite{qi2023fatezero, fruhstuck2023vive3d,lee2023soundini}. 
Another aspect involves altering the visual appearance of frames, such as transforming them into sketches or anime, without removing or replacing any objects~\cite{wang2023zero}. This closely aligns with existing literature on image style transfer in terms of their end objectives. Text-driven methods for modifying visual content have also gained considerable popularity recently.
The third category encompasses tasks related to video summarization and highlight detection. In this context, the goal is to pinpoint specific frames in the video that satisfy particular queries~\cite{narasimhan2021clip, zhang2018query, yasmin2023key}~directly linked to the video's content. Finally, the fourth category pertains to the fundamental elements of video editing, including cuts~\cite{chen2023match, pardo2022moviecuts}, transitions~\cite{shen2022autotransition, tang2018fast}, and camera placements~\cite{rao2022temporal}. Additionally, there is a wealth of research focused on video frame interpolation~\cite{kalluri2023flavr, niklaus2023splatting}~and enhancing the overall rendering quality of videos~\cite{gao2023vdpve}.

\smallskip
\noindent\textbf{Visual Transitions in Videos.}
Visual transitions in video editing play a crucial role in maintaining continuity, influencing the perception of time, and amplifying the video's ability to convey specific moods and emotions. Previous research has primarily focused on identifying transitions in videos~\cite{nam2005detection} and understanding their impact on the viewing experience~\cite{zhang2018evaluating}. Among the existing literature, the work closest to our objectives is~\cite{shen2022autotransition}, where they introduce a visual transition recommendation system. While both their work and ours address the problem of recommending visual transitions, \textit{our focus is specifically on ensuring coherence and conformity of the recommended transitions with the intended production style characteristics of the video.}

\smallskip
\noindent\textbf{Video Editing Datasets.}
Current video editing-related datasets are primarily based on classifying video shots~\cite{bain2020condensed, hassanien2017large}, visual scene recognition, detecting and identifying cuts in videos~\cite{pardo2022moviecuts}, multi-camera editing~\cite{rao2022temporal}, detecting transitions and recommending transitions~\cite{shen2022autotransition}. However, there are no existing datasets, that also give information regarding what video style category these fall under. The lack of this information has made the task of solving problems related to how style associates with the different video editing non-trivial. We, therefore, extend the Autotransition dataset~\cite{shen2022autotransition} to incorporate the video style-related labels. \\

Despite these advancements, there is a notable absence of work exploring the distinct styles of video production and their unique compositions. Our work aims to bridge this gap by examining the intersection of video editing and production style, contributing to a deeper understanding of their unique characteristics. Moreover, \textit{while previous works have primarily focused on manipulating the original visual content, we instead only concentrate on altering the manner in which different video clips are stitched together to compose the video.} 

\section{Task Definition}\label{sec:task_defn}
A video \(V\) comprises of ordered video clips represented by \( V = \{c_1, c_2, ..., c_n\} \), where \( c_i \), for $i \in \{1, 2, ...,n\} \), represents the \( i^{th} \) video clip out of a total of \( n \) video clips in \( V \). 
Now, consider the task of adapting video \( V \) to adapt to a different production style \( s \). Our objective here is to estimate the sequence of visual transitions $T = \{tr_{1}, tr_{2}, ..., tr_{n-1}\}$ that will facilitate the transformation of the video for the given desired video style $s$. Here, $tr_{k}$ for $k \in$ $\{1, ..., n-1\}$ is the transition being recommended between video clips $c_{k}$ and $c_{k+1}$ in the adapted video. $T$ can be an altered version of the transitions that \(V\) might have had before production style adaptation or an entirely new sequence. In our method, we take \(V\) and $s$ as input and generate $T$ as the output.

\section{Dataset}
\label{sec:datasets}

\subsection{AutoTransition Dataset}
\label{sec:data-autotransition}
This dataset~\cite{shen2022autotransition}~primarily consists of 35k videos with each video annotated for the set of transitions that have occurred. They provide information regarding the class of transition, the timestamp of when it begins, when it ends, duration, and type of transition. Overall, the dataset has 93 transitions. However, most of the dataset transitions are spread across 30 different visual transitions. Some of these transitions include pull-in, mix, dissolve, and black fade.

\paragraph{\textbf{AutoTransition++ Dataset.}}
\label{sec:data-save_videos}
We release a mini representative dataset of the AutoTransition dataset comprising $6k$ videos to handle computational constraints. The transition distribution within~\dataname~mirrors that of the AutoTransition dataset. Also, in our endeavor to enable the seamless adaptation of videos across various styles using transitions, autotransition presents two significant limitations. Firstly, it \textit{only} provides a \textit{single set} of transition annotations for each video, leaving a gap in the availability of transition information tailored to diverse video styles. This absence restricts its direct applicability for adaptation across different style variations. Secondly, the dataset does not disclose the video style for each entry, impeding our ability to leverage this crucial information for more effective and nuanced video adaptation.

A careful manual examination of the production styles of the videos within the dataset resulted in categorizing the videos into five main categories: vlog, influencer-centric, nature/urban-related scenes, photo slideshows, and anime. The videos in these categories differed not only in terms of content but also in the distribution of transitions. Therefore, the categorization was found reasonable to be considered as five different video styles available in the dataset.
\begin{figure}
    \centering
    \includegraphics[width=0.6\columnwidth]{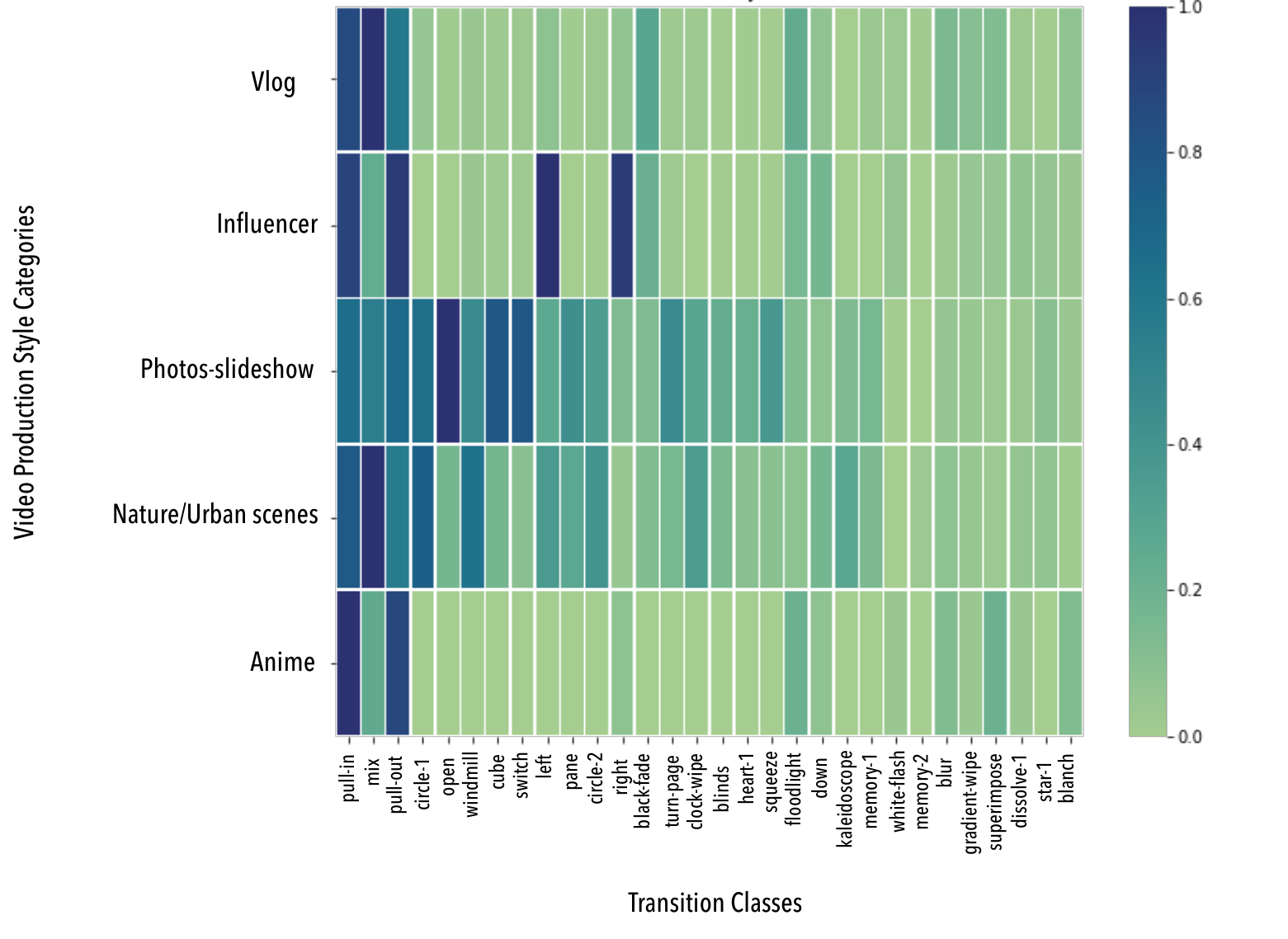}
    \caption{Bivariate distribution observed between the different styles and the visual transitions deployed across the 1379 video production style annotated videos within~\dataname.}
    \label{fig:enter-label}
\end{figure} 
We enlisted the expertise of 8 seasoned annotators who possess extensive experience in evaluating a diverse range of visual content. They were initially made to undergo a training phase where they evaluated our style categorization using a curated selection of videos spanning various styles. This not only refined their understanding of the task but also ensured uniformity and comprehension of the distinct style labels. $1379$ samples from the~\dataname~dataset were annotated with video-style labels. The distribution of visual transitions observed in the annotated samples for the different video styles has been plotted in Fig.~\ref{fig:enter-label}. Fig.~\ref{fig:enter-label} also indicates that the different styles may incorporate common transitions, potentially blurring the lines that distinguish one style from another. The discerning factor between different video styles need not be only the presence of a few unique transitions but also the frequency with which specific transitions are employed in one style as compared to the other. This behavior is observed not only among the samples annotated by us but also in video style analysis shown  by~\cite{shen2022autotransition}. However,~\cite{shen2022autotransition} have not released the video style data.

\section{~\modelname}
\label{sec:approach}
\begin{figure}[t]
    \centering
    \includegraphics[width=0.9\textwidth]{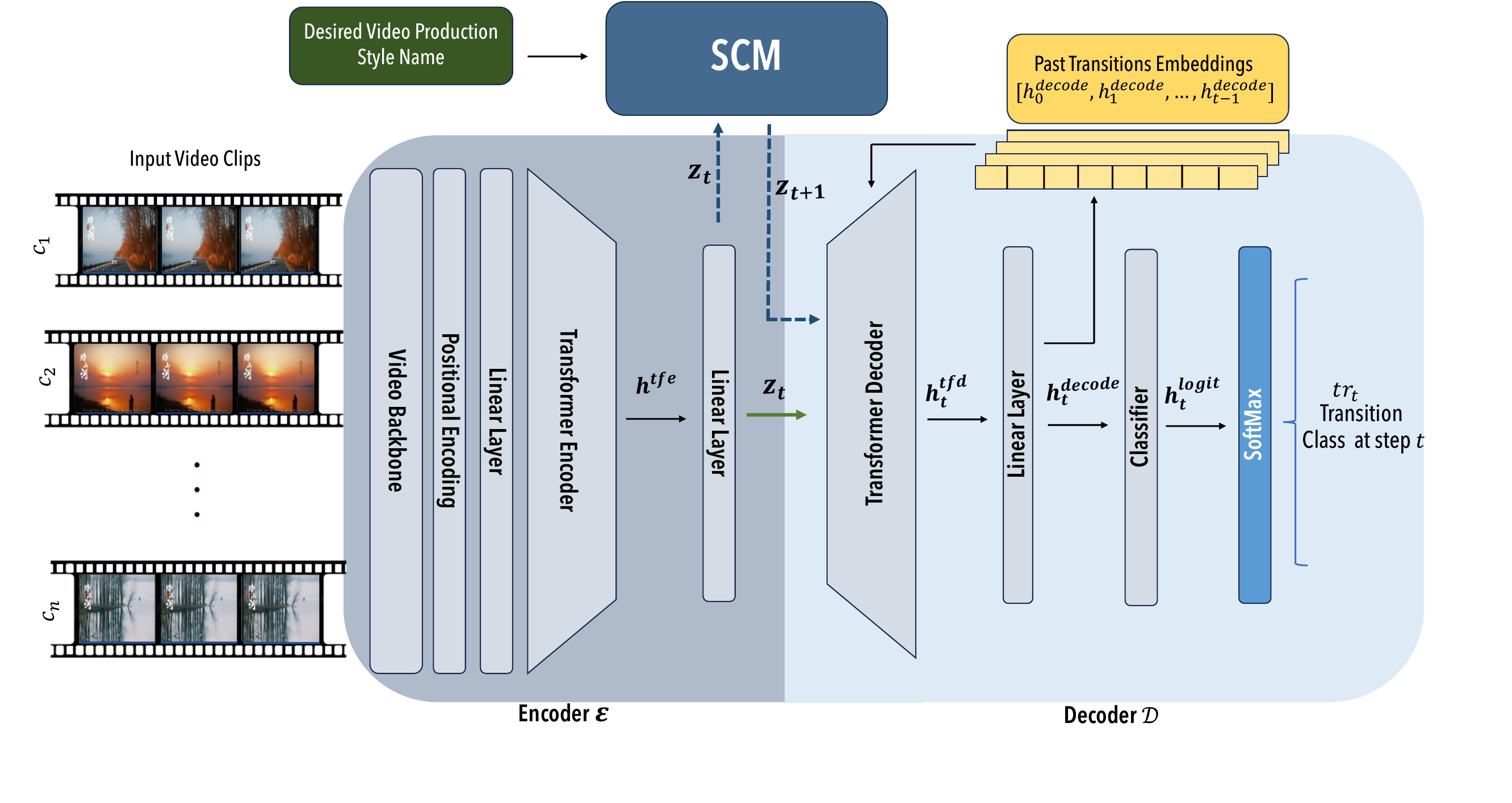}
    \caption{\modelname: Ordered clips $\{c_1, c_2,..,c_n\}$ in $V$ is fed to an Encoder $\mathcal{E}$ to obtain $z_t$. Decoder $\mathcal{D}$ takes in $z_t$ and outputs a sequence of transitions in $n-1$ steps. At each step $t$, the $\mathcal{D}$ outputs $tr_t$. The masked transformed decoder uses past transition embeddings in every step to compute $h_t^{tfd}$. $z_t$ is same (i.e., $z = z_t$) across all steps during the joint training of $\mathcal{E}$ and $\mathcal{D}$. Only components connected by $\rightarrow$ are active during training. $\mathcal{SCM}$ is an inference-time module. At inference, for every step $t$, $\mathcal{SCM}$ takes in current $z_t$ to estimate the appropriate $z_{t+1}$ for the next step to produce production style favorable transition $tr_t$. }
    \label{fig:overview}
\end{figure}
Our overarching goal is to recommend visual transitions to facilitate the adaptation of a video to a different production style. The top transitions obtained from our algorithm should ideally satisfy \textit{three} conditions. They should (a) \textit{match the video content and dynamics} (b) \textit{ensure video continuity} and (c) \textit{exhibit the desired production style characteristics}. To this end, we opt for a bottom-up approach. 
Our algorithm consists of \textit{three} main components: an encoder ($\mathcal{E})$, a decoder ($\mathcal{D}$), and a style conditioning module ($\mathcal{SCM}$). Our encoder-decoder is neural-network-based. Their combination is employed to learn a generative model of visual transitions solely from video clips. Given an input $V$ (set of ordered video clips used in input video, Sec.~\ref{sec:task_defn}), we feed it through the encoder to obtain its embedding vector $z$. $z$ is then fed to the decoder. The decoder will then generate the corresponding sequence of $n-1$ transitions that \textit{only satisfy} conditions (a) and (b). $n$ is the number of video clips in $V$. The trained encoder-decoder network is subsequently used by the most crucial part of our algorithm, the style conditioning module. It uses a gradient-based optimization approach through activation maximization~\cite{erhan2009visualizing} to update $z$. This will favorably influence the decoder outputs (i.e., the sequence of $n-1$ visual transitions) and thereby, also satisfy condition (c). 

\subsection{Pre-trained Transition and Style Embeddings}\label{sec:pretrained_transit_embed}
\begin{figure}
    \centering
    \includegraphics[width=0.7\textwidth]{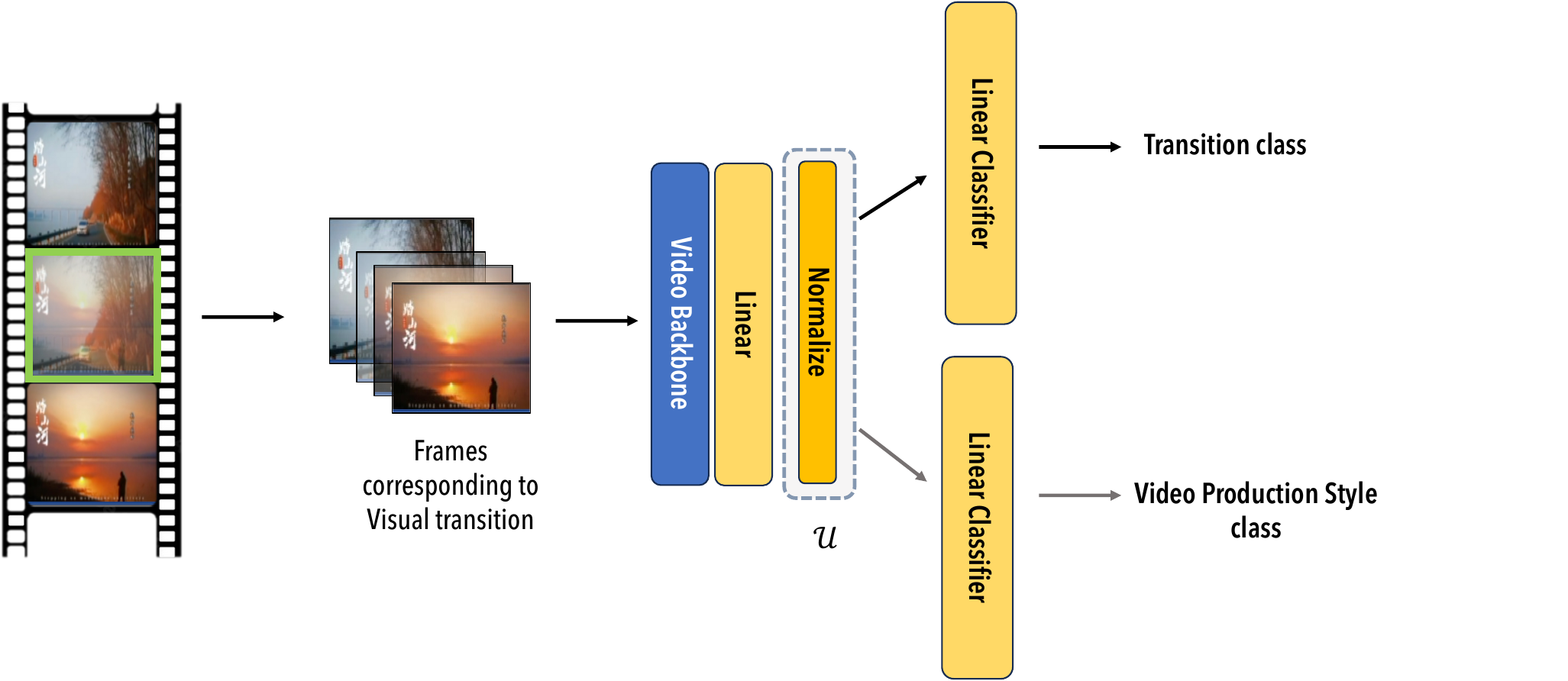}
    \caption{\textbf{$\mathcal{MLN}$}: The green box shows the region where the visual transition occurred in the video. The frames corresponding to this transition are fed to a video backbone and then through a linear layer to obtain $\mathcal{U}$ after normalization. This is then fed to linear classifiers for predicting the transition class and the video style.}
    \label{fig:cond_multitask}
\end{figure}

This step serves as a pre-processing stage in the algorithm, aiming to derive meaningful representations for distinct visual transitions and video styles. As detailed in~\cite{shen2022autotransition}, specific transitions share common visual effects, such as pull-in and pull-out. Additionally, as depicted in Fig.~\ref{fig:enter-label}, certain sets of transitions are more prevalent in videos associated with one style compared to the other. It is anticipated that the learned embeddings corresponding to these transitions will capture these characteristics. To achieve this, we employ a multitask learning-based network, $\mathcal{MLN}$ (Fig.~\ref{fig:cond_multitask}). The network optimizes on two tasks: (1) classifying visual transitions and (2) predicting the video style in which the transition will be employed. The network takes in only the video segment associated with a visual transition and obtains its visual features using a video backbone. These features are then linearly transformed and normalized to obtain a unit vector, $\mathcal{U}$. The unit vector $\mathcal{U}$ is then sent to two linear classifiers for transition class prediction and video style prediction respectively.
The embeddings corresponding to each transition in the dataset and style are sampled from $\mathcal{U}$ (Refer to appendix for details). Therefore, this will give us the set of visual transition embeddings $E_{tr} = \{e_1^{tr}, e_2^{tr}, ..., e_{N_{tr}}^{tr}\}$ where $e_i^{tr}$ for $i\in \{1, 2, ..., N_{tr}\}$ represents the embedding corresponding to the $i^{th}$ transition class. $N_{tr}$ denotes the total number of visual transition classes. With transition embeddings, we will also get the set of style embeddings, $E_{style} = \{e_1^{style}, e_2^{style}, ..., e_{N_{style}}^{style}\}$. Here, $e_i^{style}$ for $i \in \{1, 2,..., N_{style}\}$ refers to the embedding corresponding to the $i^{th}$ style and $N_{style}$ denotes the total number of video styles available.

\subsection{Encoder and Decoder}
\label{sec:appr3-module1}
Our algorithm strongly depends on the encoder $\mathcal{E}$ and decoder $\mathcal{D}$. Their collaborative operation to generate visual transitions using clips forms the core of~\modelname. The transitions obtained using $\mathcal{E}$ and $\mathcal{D}$ should take into account not only the visual content but also the temporal dependency between transitions to ensure video continuity. 
Taking this into consideration, we employ a transformer encoder-decoder-based architecture. The input is a sequence of video clips in the order they appear in the input video. The output is a sequence of transitions \textit{satisfying} transition conditions (a) and (b). 

We begin by uniformly sampling $w$ frames from each video clip in $V$. These are then fed into a video backbone following standard practices~\cite{shen2022autotransition} to obtain an embedding $e^{V}$. In our experiments, we use SlowFast~\cite{feichtenhofer2019slowfast}~as our video backbone. Before sending the visual features of the video clips to the transformer encoder, learnable position embeddings are element-wisely added to them. This is to encourage the model to consider the sequence of the video clips in the input video into consideration while recommending the sequence of transitions. The transformer encoder outputs $h^{tfe}$ that is fed to a fully connected layer to obtain $z$. Therefore, $\mathcal{E}(V) = z$.
To ensure video continuity, the decoder $\mathcal{D}$ is designed to recommend one transition at a time, relying on previously generated transitions. It uses a masked transformer decoder using a masked multi-head self-attention on the target transition sequence. The attention for each transition is restricted to cover only those preceding it in the sequence by using a triangular mask. This ensures that $\mathcal{D}$'s transition recommendation is causal and therefore, usable during inference time when the full target transition sequence is not available. For $n$ video clips in $V$, $\mathcal{D}$ recommends transitions across $n-1$ steps. At each step $t$, $z$ along with transitions observed till step $t-1$ (obtained after masking) fed to the transformer decoder outputs $h_t^{tfd}$. Subsequently, $h_t^{decode}$ is fed into another linear layer to produce $h_t^{logit}$.
To train the encoder-decoder part of the algorithm, we make use of the following loss function:
\begin{equation}
    L_V = L_{classification} + \lambda L_{masked\_triplet}
\end{equation}
where $L_{classification}$ corresponds to the cross entropy loss computed between $h_t^{logit}$ and $y_t^{trans}$ (the ground truth transition classes for step t) across all time steps. $\lambda$ is a scalar hyperparameter. $L_{masked\_triplet}$ is the masked triplet loss computed as:
\begin{equation}\label{eqn:masked_triplet_loss}
    L_{masked\_triplet} = \frac{1}{N_{tr}-1}\sum_{i\neq gt, i\in1,...,N_{tr}} \mathcal{T}(h_t^{decode}, e_{gt}^{tr}, e_i^{tr})
\end{equation}
where $e_{gt}^{tr}$ represents the transition embedding corresponding to the ground truth transition class and $e_i^{tr}$ represents the transition embedding corresponding to the $i^{th}$ class of transitions, except ground truth class. $\mathcal{T}$ calculates the triplet margin loss for each triplet ($h_t^{decode}, e_{gt}^{tr}, e_i^{tr}$)
\begin{equation}
    \mathcal{T}(a, p, n) = max(\phi(a, p) - \phi(a, n) + m, 0)
\end{equation}
$m$ is the soft margin, $a, p, n$ are anchor, positive sample and negative samples respectively. $N_{tr}$ is the total number of visual transition classes available. $\phi(x, y)$ denotes the similarity metric between $x$ and $y$. In this case, it is taken to be the dot product between the two. 
We can define $\phi$ therefore as, 
\begin{equation}
    \phi (x, y) = <x, y>
\end{equation}

\subsection{Style Conditioning Module}\label{sec:appr3-module2}
$\mathcal{E}$-$\mathcal{D}$ network construction and training objective will only enable attainment of visually and temporally coherent transitions. 
Incorporating video production style information within the $\mathcal{E}$-$\mathcal{D}$ network itself can be complex due to the absence of task-appropriate datasets (as discussed in Sec.~\ref{sec:intro} and~\ref{sec:datasets}). Therefore, we propose a \textit{style conditioning module}($\mathcal{SCM}$) to alter (if needed) the transition choices made by the decoder above to align them with the requirements of the desired video style. It is designed to eliminate the need to train $\mathcal{E}$-$\mathcal{D}$ for each style separately or have access to transition annotations for each style for each video sample. $\mathcal{SCM}$ is activated after training the encoder-decoder network during inference time. Its operational principle relies on the fundamental recognition that the outputs derived from the decoder are driven by $z$. Therefore, this implies that any alteration to $z$ will change the decoder outputs. We leverage this simple yet crucial property to achieve the purpose of this module in~\modelname. We implement the style conditioning module by employing a gradient-based optimization strategy, guided by \textit{activation maximization(AM)}~\cite{erhan2009visualizing}~to update $z$. The update is designed to ensure that the transition sequence generated by the decoder $\mathcal{D}$ exhibits the desired production style characteristics.
We execute $\mathcal{SCM}$ at every decoding step during inference. Therefore, let $z_t$ be the input to $\mathcal{D}$ at step $t$. 
In our case, AM refers to the process of using backpropagation through the decoder weights to identify a permutation of $z_t$ to bring about a desired change in the output of the decoder at step $t+1$. This therefore requires us to establish one or more distinct differentiable loss functions that can guide the optimization of $z_t$. These loss functions are equivalent to setting constraints that the updated $z_t$ should try to satisfy. In our case, we have two loss functions.\\
\noindent\textbf{Embedding Loss $(\mathcal{L_E})$.} This ensures that $z_t$ is updated in a direction that results in the decoder generating a transition sequence that facilitates the adaptation of the input video to the intended style.
\begin{equation}
    \mathcal{L_E} = 1 - \sigma_{cos}(e_\mu, e_k^{style})
\end{equation}
where $\sigma_{cos} (a,b)$ refers to the cosine similarity computed between two \textit{d}-dimensional vectors a, b $\in R^d$, i.e., 
\begin{equation}
  \sigma_{cos} (a, b) = \frac{<a, b>}{||a||_2 . ||b||_2}  
\end{equation}
Additionally $e_k^{style}$ is the embedding corresponding to the $k^{th}$ video style among $N_{style}$ video styles (Refer Sec.~\ref{sec:pretrained_transit_embed}) and we define $e_\mu$ as the average of embeddings corresponding to transitions that have been observed till decoder step $t$ for the video. 
\begin{equation}
  e_\mu = \sum_{i=1}^{t} h_i^{decode}  
\end{equation}
\noindent\textbf{Reconstruction Loss$(\mathcal{L_R})$.} This ensures that there isn't a significant deviation in the values of $z_t$ that can compromise the integrity of the information derived from the input video. To compute this, we trained a decoder $\mathcal{D}_\psi$ that takes in $z_t$ to retrieve back $h^{tfe}$. Referring to Sec.~\ref{sec:appr3-module1},  $h^{tfe}$ was obtained after encoding the input and $z_t$ after processing it further. Therefore, even after the alteration of $z_t$, we can be sure that the input information has still been preserved in $z$ if we can reliably reconstruct $h^{tfe}$. More details about $\mathcal{D}_\psi$  can be found in the appendix. Essentially, let 
\begin{equation}
    \mathcal{D}_\psi (z_t) = \hat h^{tfe}
\end{equation}
Therefore, the reconstruction loss $\mathcal{L_R}$ can be computed as
\begin{equation}
    \mathcal{L_R}(z) = |\hat h^{tfe} - h^{tfe}|_1
\end{equation}
$h^{tfe}$ is not updated in this process. Therefore $\forall$ decoding step $t=$ 1 to $n-1$, $z_t$ will be optimized to reconstruct the same $h^{tfe}$.
The overall loss function used for the optimization of $z_t$ 
\begin{equation}
    \mathcal{L}(z_t) = \alpha_\mathcal{E} \mathcal{L_E} + \alpha_\mathcal{R} \mathcal{L_R}
\end{equation}
where $\alpha_\mathcal{E}$ and $\alpha_\mathcal{R}$ are the step sizes corresponding to $\mathcal{L_E}$ and $\mathcal{L_R}$ respectively. $z_k$ is initially assigned to be equal to $z_t$. The following optimization process is then run for $5000$ iterations with a learning rate $\beta$ to finally obtain $z_{t+1}$ (Ref. Fig~\ref{fig:overview}).:
\begin{equation}
    z_{k+1} \leftarrow z_k - \beta \nabla \mathcal{L}(z_k)
\end{equation}

\noindent\textit{Finetuning the transition sequence:} After $n-1$ decoder steps, we obtain the preliminary sequence of transitions T that can enable video style adaptation. Optionally, we can finetune this sequence further if needed without affecting any parameters or disrupting any required conditions drastically. We take inspiration from RRT~\cite{karaman2011anytime}, a path-planning algorithm to do so. We consider the sequence of $h_t^{decode}$ for $t \in \{1, 2, ...n-1\}$ that was generated in the process to obtain $tr_t$. By the training construction of the decoder, $tr_t$ corresponds to the transition class whose pre-trained embedding is closest to $h_t^{decode}$. For RRT instead, we consider its $K$ nearest pre-trained embeddings using a distance metric. $K$ value should not be too high or too low. A high $K$ value can change $T$ more than we need and a very low $K$ might not make much difference to $T$. $K$ denotes the extent of exploration we wish to do. Each of the $K$ transitions is then iteratively evaluated to compute $e_\mu$ that is closest to the desired video style embedding, i.e., $e_k^{style}$. 

\section{Experiments and Results} \label{sec:exp_results}
We first discuss implementation details, the metrics used, and baselines compared with in Sec.~\ref{sec:implement}. Our algorithm is a bottom-up approach to obtain visual transitions that can facilitate the adaptation of a video to a different production style. Each component plays a crucial role in building the entire algorithm. Sec~\ref{sec:expr_encode_decode}~analyzes each of their performance in detail.
\subsection{Implementation Details}\label{sec:implement}
\noindent\textbf{Model Details.}
Similar to AutoTransition~\cite{shen2022autotransition}, we have employed SlowFast 8x8 as the video backbone to extract visual features. This is used in the encoder component as well as for both obtaining pre-trained transition and style embeddings. $\mathcal{E}$ and $\mathcal{D}$ each use $2$ transformer layers for encoding and decoding with $d_{model} = 512$ and $n_{head} = 8$. 

\smallskip
\noindent\textbf{Data Pre-processing and Training Details.} The maximum number of ordered clips considered in a video is user defined. For experiment purposes, we have considered it as $8$. For both, obtaining pre-trained transition embeddings and transition sequence recommendations, we sample $16$ frames with an image size of $224$ x $224$ from each video clip in $V$. The former uses a batch size of $64$ while the latter uses $16$. Both training processes are run for a total of $60$ epochs. The model parameters of the last epoch are used to obtain the embeddings corresponding to the $N_{tr} = 30$ transition classes and $N_{style} = 5$ video production style classes from $\mathcal{MLN}$(Fig.~\ref{fig:cond_multitask}). We train all our networks using the Adam optimizer~\cite{kingma2014adam} with a learning rate 1e-5. The margin $m$ in triplet loss is set to $0.5$. We run our experiments on the~\dataname~dataset. For training $\mathcal{E}$ and $\mathcal{D}$, we use $80\%$ of the data for training, validate the performance on $10\%$ of the data, and test on the remaining $10\%$ of the data. We train all models on $8$ A$5000$ GPUs. In $\mathcal{SCM}$, we use an SGD optimizer~\cite{robbins1951stochastic}~for optimization and a learning rate of $0.1$, $\alpha_\mathcal{E}$ and $\alpha_\mathcal{R} = 1$. All codes were implemented using Pytorch~\cite{imambi2021pytorch}.

\smallskip
\noindent\textbf{Metrics and Baselines.}\label{sec:expr_evaln_metrics}
For a fair evaluation of \textit{only} our encoder-decoder model and pre-trained embeddings, we consider the metrics used by the only other relevant baseline, AutoTransition. These include Recall@K where K$\in\{1, 5\}$ and Mean Rank. The evaluation is done for every transition in the sequence. For $\mathcal{SCM}$, we consider cosine similarity, which informs us about how relevant a method's recommended transitions are for the video production style we wish to achieve. 

\begin{figure}[ht]
  \begin{subfigure}{0.5\textwidth}
    \centering
    \resizebox{\textwidth}{!}{
    \begin{tabular}{lcc}
        \toprule
        \multirow{1}{*}{\textbf{Transition}} & \multicolumn{2}{c}{\textbf{Evaluation Metrics}}\\
        \cmidrule{2-3}
        \textbf{Embedding}& \small{Recall@1$\uparrow$} & \small{Recall@5$\uparrow$}\\
        \midrule
        Transition Classifier~\cite{shen2022autotransition} & 97.85 & 99.78\\
        $\mathcal{MLN}$ Task-1 & 96.67 & 98.77\\
        $\mathcal{MLN}$ Task-2 & 93.45 & 99.54 \\
        \bottomrule
    \end{tabular}
    }
    \caption{}
    \label{tab:my_table}
  \end{subfigure}
    \begin{subfigure}{0.5\textwidth}
    \centering
    \includegraphics[width=\linewidth]{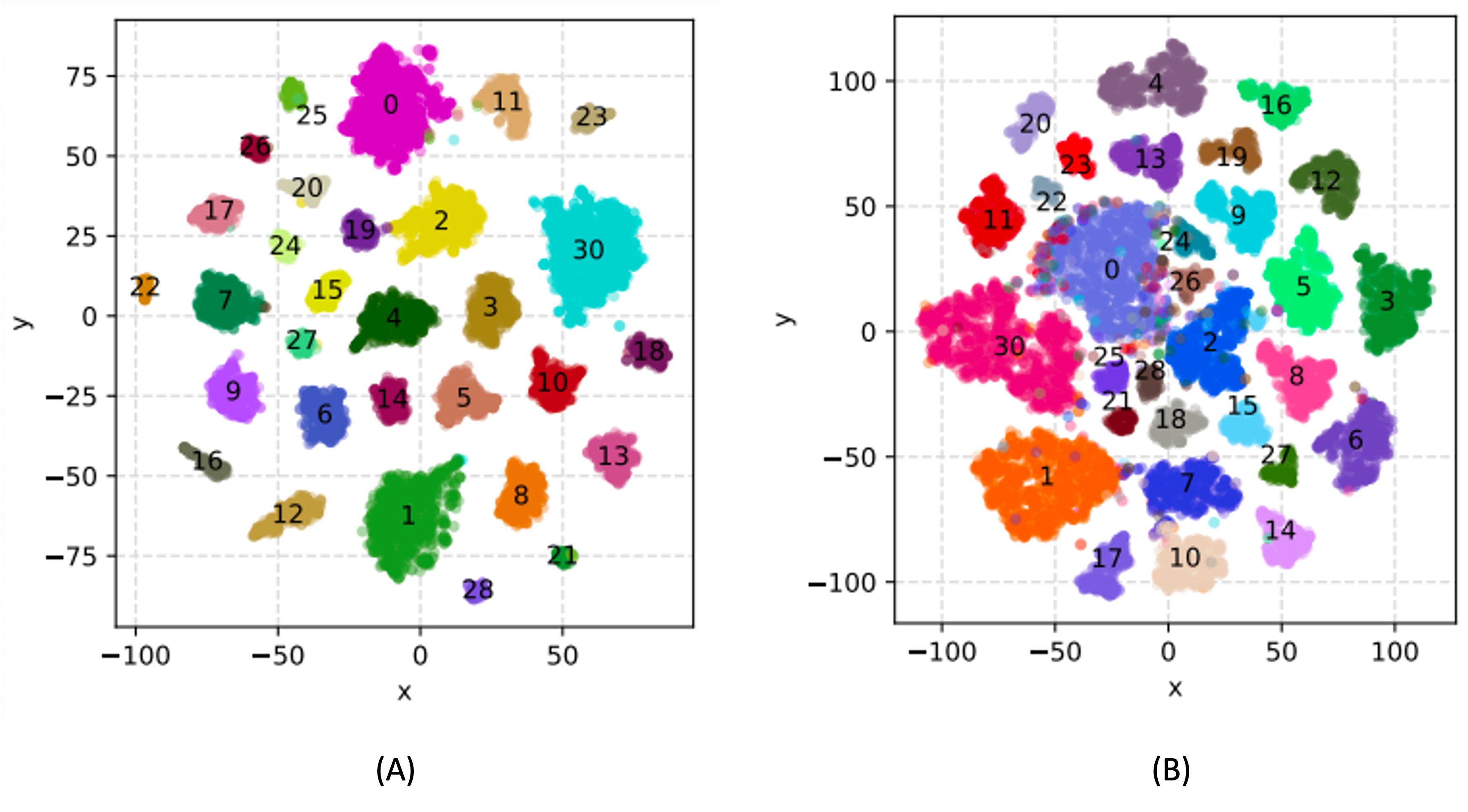}
    \caption{}
    \label{fig:my_figure}
  \end{subfigure}%
  \caption{\textbf{(a)} Compares transition classification results between~\cite{shen2022autotransition}'s classifier and MLN. MLN Task-1 is for Transition classification, and Task-2 is for video production style classification. \textbf{(b)} Displays t-SNE visualizations of the transition embeddings. Fig. (A) depicts embeddings from~\cite{shen2022autotransition}'s method, while (B) represents embeddings from MLN.}
  \label{fig_table_combination}
  \smallskip
\end{figure}
\subsection{Comparisons and Analysis}\label{sec:expr_encode_decode}
\noindent\textbf{Pre-Trained Transition and Style Embeddings.} Fig.\ref{fig_table_combination} provides insights into the quality of pre-trained transition embeddings used for training $\mathcal{E}$ and $\mathcal{D}$, and style embeddings for $\mathcal{SCM}$. $\mathcal{MLN}$ differs from~\cite{shen2022autotransition}'s classifier in its ability to obtain transition embeddings that obeyed both visual and style similarities. Given the relatively smaller number of constraints, it is expected that the transition embeddings in the case of~\cite{shen2022autotransition}~ will be comparatively more distinctly placed (as observed in Fig.\ref{fig:my_figure}) as compared to that obtained using $\mathcal{MLN}$. This comparison aims to highlight that the additional style constraint in $\mathcal{MLN}$ doesn't significantly compromise transition classification accuracy while maintaining high video style classification. Therefore, the differences between different transition and style classes have been captured by $\mathcal{MLN}$ reliably.

\smallskip
\noindent\textbf{Encoder-Decoder.}
In this work, we have limited our scope to understanding visual content in videos and visual transitions. We, therefore, compare AutoTransition's performance when using only visual inputs. Table~\ref{res:encode_decode} shows the results obtained from our experiments. Three notable observations emerge. Firstly, both methods mostly demonstrate superior performance with pre-trained transition embeddings compared to randomly initialized embeddings. This underscores the effectiveness of leveraging pre-trained embeddings for better mapping of the visual space to the transition space.
Secondly, our encoder-decoder model consistently outperforms AutoTransition across all metrics, regardless of the chosen type of embedding. This suggests that capturing temporal relations between transitions and visual content leads to more accurate recommendations for transitions. Thirdly, examining our model's performance with transition embeddings from~\cite{shen2022autotransition}versus those from $\mathcal{MLN}$ reveals intriguing insights. Although the difference is subtle, it's worth noting that Recall@1 is higher when using\cite{shen2022autotransition}'s embeddings, whereas Recall@5 exhibits the opposite trend. This can be attributed to the distribution of transitions in the embedding space. Fig.\ref{fig:my_figure}(A) demonstrates better separation between transitions, explaining the higher precision of our encoder-decoder trained on\cite{shen2022autotransition}. On the other hand, Fig.~\ref{fig:my_figure}(B) illustrates closer proximity between transitions, elucidating the higher Recall@5 when trained using $\mathcal{MLN}$

\smallskip
\noindent\textbf{Ablation Experiments on Encoder-Decoder Losses.} Table~\ref{tab:loss_ablation} summarizes results from an ablation experiment on the two loss functions used. When using $L_{classification}$ alone, we observe the highest Recall@5 but the lowest Recall@1. Conversely, using only $L_{masked\_triplet}$ shows the opposite trend. The combination of both losses provides the most balanced results. This behavior can be attributed to the specific strengths of each loss function, with $L_{classification}$ emphasizing broader classification accuracy and $L_{masked\_triplet}$ focusing on intricate relationships within the data. The combination of both losses allows for a balanced optimization, leveraging the strengths of each to achieve a more comprehensive performance across both metrics.
\begin{table*}[t]
    \centering
    \caption{\textbf{Quantitative Results.} We show the quantitative results of our encoder-decoder model compared to AutoTransition~\cite{shen2022autotransition} (only visual component included), our baseline.} 
    \label{tab:ablation}
    \resizebox{\textwidth}{!}{%
    \begin{tabular}{clccccc}
    \toprule
    \multirow{2}{*}{\textbf{Transition}} & \multirow{2}{*}{\textbf{Method}} & \multicolumn{3}{c}{\textbf{Evaluation Metrics}} \\
    \cmidrule{3-5}
    \textbf{Embedding}& & \small{Recall@1$\uparrow$} & \small{Recall@5$\uparrow$} & \small{Mean Rank$\downarrow$} \\
    \midrule
    \multirow{2}{*}{Random} & AutoTransition~\cite{shen2022autotransition} & 15.24 & 42.00 & 9.1 \\
    \cmidrule{2-5}
    &\textbf{~\modelname($\mathcal{E}+\mathcal{D}$)} & 18.43 & 47.39 & 8.19\\
    \midrule
    \multirow{2}{*}{Transition Embedding~\cite{shen2022autotransition} (Pre-Trained)} & AutoTransition~\cite{shen2022autotransition}& 16.93 & 44.58 & 8.82\\
    \cmidrule{2-5}
    &\textbf{~\modelname($\mathcal{E}+\mathcal{D}$)} & 23.24 & 56.06 & 6.77\\
    \midrule
    \multirow{2}{*}{$\mathcal{MLN}$ Transition} & AutoTransition~\cite{shen2022autotransition}& 12.12 & 36.66 & 10.48\\
    \cmidrule{2-5}
    Embedding (Pre-Trained)&\textbf{~\modelname($\mathcal{E}+\mathcal{D}$)} & 22.59 & 61.03 & 5.90\\
    \bottomrule
    \end{tabular}
    }
    \label{res:encode_decode}
\end{table*}

\begin{table*}[t]
    \centering
    \caption{\textbf{Loss Ablation Experiments.} We show ablation experiments on the loss functions used to train our encoder-decoder model. Bold denotes \textbf{best} while underline denotes \underline{second-best}}
    \label{tab:ablation}
    \begin{tabular}{cccccc}
    \toprule
    \multirow{2}{*}{\textbf{Loss}}& \multicolumn{3}{c}{\textbf{Evaluation Metrics}} \\
    \cmidrule{2-4}
    \textbf{Function}& \small{Recall@1$\uparrow$} & \small{Recall@5$\uparrow$} & \small{Mean Rank$\downarrow$} \\
    \midrule 
    \multirow{1}{*}{$L_{classification}$} & 11.63 & \textbf{69.00} & 11.41 \\
    \midrule
    \multirow{1}{*}{$L_{masked\_triplet}$}& \underline{16.89} & 45.79 & \underline{8.57}\\
    \midrule
    \multirow{1}{*}{$L_{classification} + L_{masked\_triplet}$} & \textbf{22.59} & \underline{61.03} & \textbf{5.90}\\
    \bottomrule
    \end{tabular}
    \label{tab:loss_ablation}
\end{table*}

\smallskip
\noindent\textbf{Style conditioning Module ($\mathcal{SCM}$).}
To evaluate $\mathcal{SCM}$, we randomly selected $100$ videos from the test dataset, each comprising more than $6$ video clips. For each of the production styles defined in~\dataname~dataset, we computed the cosine-similarity between $e_\mu$, i.e. the mean embedding of all the transitions predicted for a video and the desired video production style embedding. Table~\ref{res:SCM} shows the mean of the cosine similarity scores obtained across the $100$ samples for each of the baseline methods. Higher similarity scores are better. Apart from \textit{Anime}, we can observe that $\mathcal{SCM}$ improves the suitability of transitions for adapting to the desired production style.  
\begin{table*}[t]
    \centering
    \caption{\textbf{Qualitative Results - SCM.} We show the quantitative results of our algorithm to recommend transitions to enable the adaptation of video production styles with baselines. Bold denotes \textbf{best} while underline denotes \underline{second-best}. The higher the similarity better it is.} 
    \label{tab:ablation}
    \resizebox{\textwidth}{!}{%
    \begin{tabular}{clccccc}
    \toprule
    \multirow{2}{*}{\textbf{Method}} & \multicolumn{5}{c}{\textbf{Video Production Styles}} \\
    \cmidrule{3-7}
    & & Vlog & Anime & Influencer & Photos & Nature \\
    \midrule
    \multirow{1}{*}{AutoTransition~\cite{shen2022autotransition}} & & -0.007 &\textbf{ 0.0308} & -0.089 & -0.05 & -0.025 \\
    \multirow{1}{*}{~\modelname ($\mathcal{E}$ \& $\mathcal{D}$) }& & -0.01 & -0.035 & -0.022 & -0.018 & 0.013 \\
    \multirow{1}{*}{~\modelname ($\mathcal{E}$ \& $\mathcal{D}$, $\mathcal{SCM}$ w/o RRT) }& & \underline{0.018} & \underline{0.0194} & \underline{0.0102} & \underline{0.0211} & \underline{0.0496} \\
    \multirow{1}{*}{~\modelname ($\mathcal{E}$ \& $\mathcal{D}$, $\mathcal{SCM}$ w/ RRT) }& & \textbf{0.188} & -0.001 & \textbf{0.227} & \textbf{0.7286} & \textbf{0.356} \\
    \bottomrule
    \end{tabular}
    }
    \label{res:SCM}
\end{table*}

\section{Conclusion, Limitations and Future Work}
We present~\modelname, a bottom-up approach for recommending visual transitions that adapt videos to different production styles, such as documentaries, feature films, and dramas. Our 3-component algorithm inputs a video and desired production style class to output a sequence of recommended transitions. The encoder-decoder model maps the vision space to the transition space, ensuring visually and temporally consistent recommendations. The style conditioning module uses activation maximization to adjust the latent embedding, enabling the decoder to produce style-specific transitions. We demonstrate our results on~\dataname, a mini version of the AutoTransition dataset with 6k videos. While this work focuses on visual transitions, future research will explore other editing elements like camera motion, cuts, and audio. Nevertheless, our current work lays a crucial foundation for understanding video production styles and their connections to editing elements.
\clearpage  

\section*{Acknowledgements}
This work was done by Pooja as part of an internship with Dolby Laboratories. 

%
%
\bibliographystyle{splncs04}
\bibliography{main}
\include{supplementary}
\end{document}

%% file: macros.tex
\usepackage{array}
\usepackage{colortbl}
\usepackage{xcolor}
\usepackage{multirow, makecell}
\usepackage{subcaption}
\usepackage{booktabs}
\usepackage{amsmath}
\usepackage{amssymb}
\usepackage{amsfonts}
\usepackage{enumitem}
\usepackage{dsfont}

\newcommand{\modelname}{V-Trans4Style}
\newcommand{\dataname}{AutoTransition++}

%% file: supplementary.tex
\clearpage
\setcounter{page}{1}
\title{\modelname: Visual Transition Recommendation for Video-Production Style Adaptation \\ ---Appendix---}
\titlerunning{V-Trans4Style}
\author{Pooja Guhan\inst{1}\orcidlink{0000-0003-1551-8163} \and
Tsung-Wei Huang\inst{2}\orcidlink{0000-0002-1478-2678} \and
Guan-Ming Su\inst{2}\orcidlink{0000−0002−3118−5904}\and 
\\Subhadra Gopalakrishnan\inst{2}\orcidlink{0009-0007-7878-6858}\and
Dinesh Manocha\inst{1}\orcidlink{0000-0001-7047-9801}}

\authorrunning{Guhan, Pooja et al.}

\institute{University of Maryland, College Park MD 20740, USA \and
Dolby Laboratories, Sunnyvale CA 94085, USA}

\maketitle

\section{\dataname~Details}
\begin{figure}
    \centering
    \includegraphics[width=\linewidth]{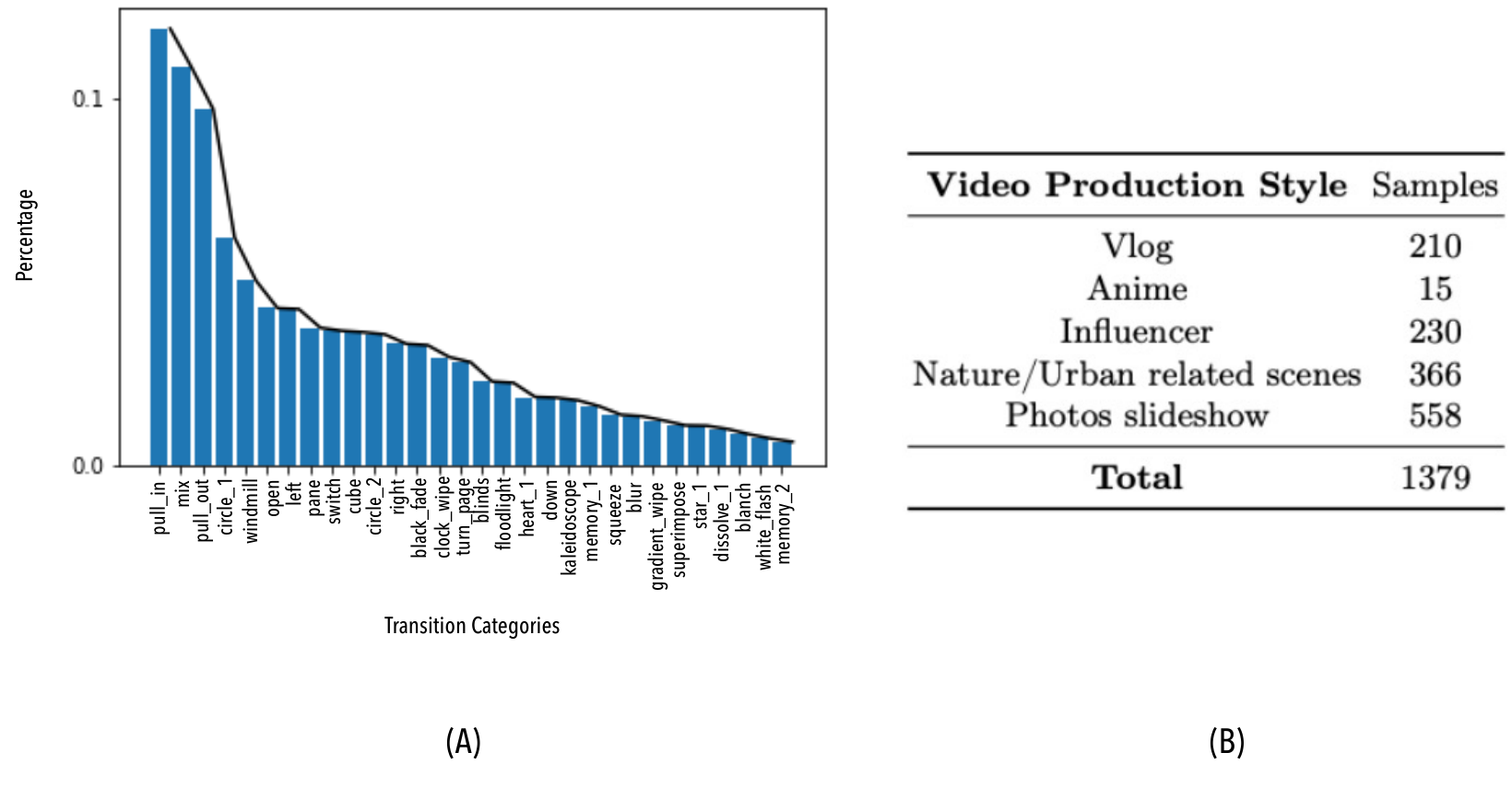}
    \caption{(A) shows the transition distribution in~\dataname~dataset. (B) shows the distribution of video production style labels available.}
    \label{fig:autotrans_mini_distr}
\end{figure}

We have collected and showcased our proposed method of recommending transitions based on the desired video production style on~\dataname, a mini representative version of AutoTransition dataset. It consists of $6k$ videos. The distribution of transitions in~\dataname~has been shown in Fig.~\ref{fig:autotrans_mini_distr}.A.  $1379$ videos from these have been additionally annotated so far for video production styles. We define the production styles considered as follows:
\begin{enumerate}
    \item Vlog: A vlog is a video format that primarily centers around documenting and sharing aspects of human experiences, including life, stories, or day-to-day events. The content often revolves around the human subject, offering viewers a personal and relatable glimpse into their world. Whether recounting personal anecdotes, showcasing daily routines, or sharing unique experiences, the essence of a vlog lies in its ability to intimately connect the audience with the broader spectrum of human perspectives and lives.
    \item Anime: These typically consist of compiled segments or scenes from anime shows.
    \item Influencer: These typically revolve around content related to fashion, lifestyle, and personal experiences. This may encompass elements such as photoshoots, selfies, and videos that highlight the influencer's unique style and interests.
    \item Nature/Urban Scenes: These feature content centered around natural environments, encompassing landscapes, wildlife, and even urban settings. These videos may showcase the beauty of nature, including animals, and provide glimpses into the diverse landscapes of urban areas.
    \item Photo Slideshow:  These videos cover a range of themes, including images commemorating significant life moments, compiling cherished memories (from outings or important events), promoting a brand, or offering glimpses into a forthcoming drama series by showcasing scenes captured during production. These videos may be a seamless slideshow, weaving the images together to present a cohesive narrative of a specific day or event. They can also feature slideshows of inanimate objects, food, and subjects beyond personal experiences.
\end{enumerate}
These definitions were agreed upon between all annotators before they started with the annotation process. While the AutoTransition dataset~\cite{shen2022autotransition}~suggests the availability of video style-related information, we still believe that the availability of AutoTransition++ is useful due primarily due the following reasons:
\begin{enumerate}
    \item The style information claimed by~\cite{shen2022autotransition}~is not publicly available. 
    \item The styles discussed in~\cite{shen2022autotransition} seems to focus more on creating visually appeal and emotions. The labels introduced by us on the other hand are more to do with the format and content delivery than a specific visual or emotional style. They describe the type of content and the way it is presented. We believe this would be a useful addition to the dataset and hence the creation of AutoTransition++.  
\end{enumerate}
\paragraph{Ethical Considerations: } The data collection process does not include any personal, private or sensitive information, and was deemed exempt from an ethics review. The videos being annotated for video production styles are part of an existing dataset AutoTransition. Additionally, the annotators were instructed to flag any video with harmful or offensive content. \\

The dataset can be accessed on our project webpage: \url{https://gamma.umd.edu/v-trans4style/}
\section{Model Details}
In this section, we provide additional details for two auxiliary models mentioned in our work to develop~\modelname. These include namely,$\mathcal{MLN}$ (Sec.~\ref{sec:aux_mln}) and $\mathcal{D}_\psi$ (Sec.~\ref{sec:aux_dpsi}).
\subsection{$\mathcal{MLN}$}\label{sec:aux_mln}
The network, $\mathcal{MLN}$ shown in Fig.~\ref{fig:cond_multitask}(in main paper) was used to obtain the pre-trained embeddings for different transitions and video production styles. The following loss function was used for training purposes:

\begin{equation}\label{eqn:cond_ml}
    L_{mtl} = L_{TC} + \lambda L_{VPC},
\end{equation}
$L_{TC}$ corresponds to the loss function for transition classification and $L_{VPC}$ corresponds to the loss function for video production style classification. Both $L_{TC}$ and $L_{VPC}$ are cross-entropy ($CE$) loss functions. Training this network, however, poses a challenge from the data perspective. More specifically, it relates to the uneven number of labels available for each task. In the context of our problem, all samples within the~\dataname~dataset can be used for training the transition classifier. However, not all of these samples have labels corresponding to video production style too. Therefore, this imbalance in labeled samples across both the tasks (i.e., transition and video production style classification) can present challenges when training the multitask network $\mathcal{MLN}$. We, therefore, adopt \textit{selective task activation} based training mechanism to maximize learning by using the full potential of all our data samples. This implies that we activate the branch corresponding to video production style classification only for inputs containing labels for video production style. In other cases, only loss for transition classification is computed. Therefore the loss function can be modified as follows:
\begin{equation}
    L_{mtl} = \begin{cases} 
L_{TC} + \lambda L_{VPC} & \text{if } \hat{y}_{VPC} \text{ exists}, \\
L_{TC} & \text{otherwise} \\
\end{cases}
\end{equation}
$\hat y_{VPC}$ refers to the video production style label for the sample. 
\paragraph{Training Details: } Dataset split mentioned in Sec.~\ref{sec:implement} remains the same for training $\mathcal{MLN}$. We used a batch size of $64$ and ran training for $30$ epochs. We used adam optimization with an initial learning rate of 1e-3. 
\subsection{$\mathcal{D}_\psi$}\label{sec:aux_dpsi}
This component is introduced to calculate the reconstruction loss as defined in Section \ref{sec:appr3-module2}. In this context, $\mathcal{D_\psi}$ comprises two linear layers designed to take input $z$ and learn to produce $h^{tfe}$. During training, the encoder $\mathcal{E}$ remains fixed, and the decoder $\mathcal{D}$ is not utilized. When presented with an input, $\mathcal{E}$ generates $h^{tfe}$ and subsequently derives $z$. This $z$ is then fed into $\mathcal{D}_\psi$ to $\hat h^{tfe}$, i.e., the reconstructed $h^{tfe}$. We compute L1 loss between $h^{tfe}$ and $\hat h^{tfe}$ for training $\mathcal{D}_\psi$. 
\begin{figure}
    \centering
    \includegraphics[width=0.7\linewidth]{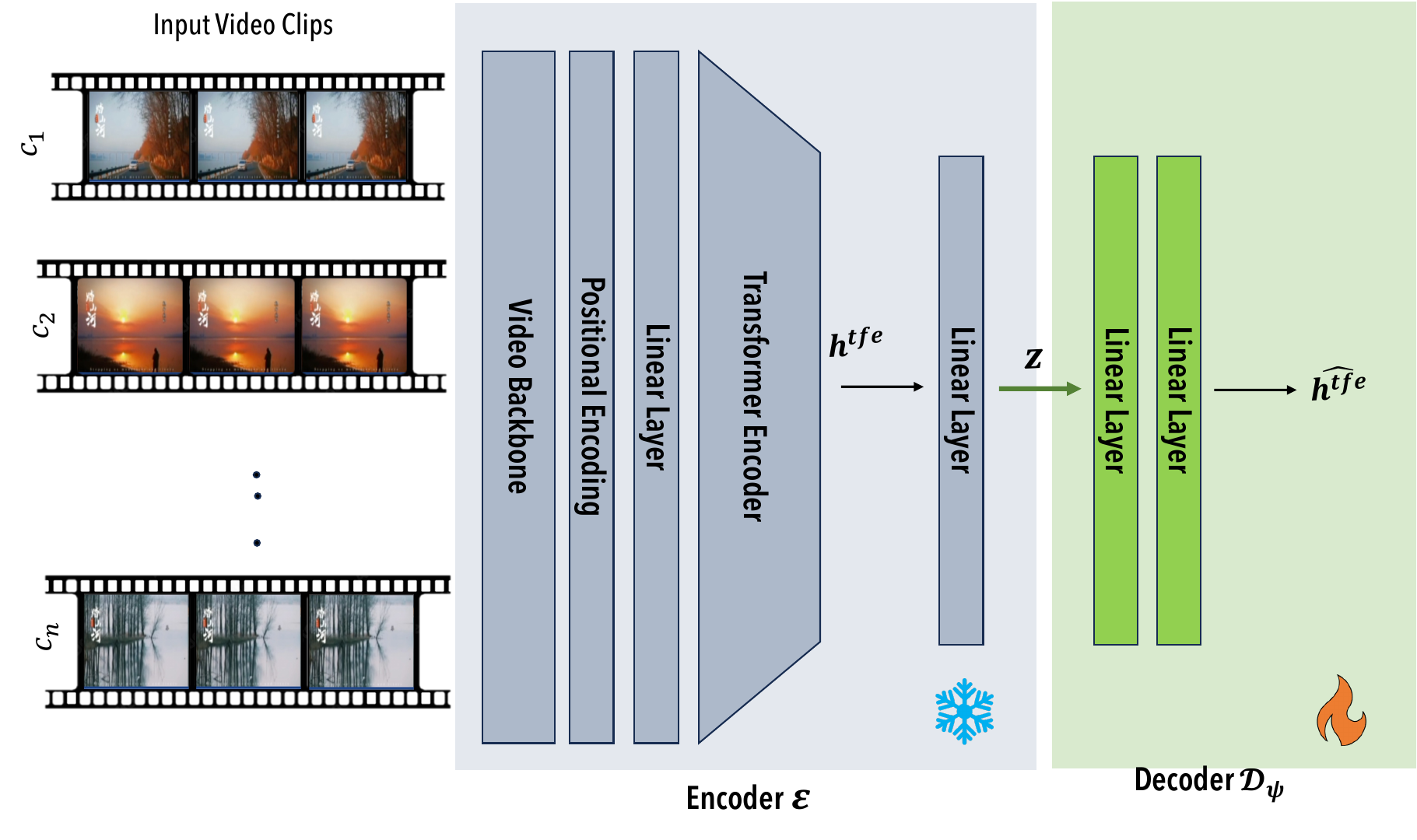}
    \caption{$\mathcal{D}_\psi$ is used as a model capable of reconstructing the encoder feature vector $h^{tfe}$. This property is used in the development of the reconstruction loss mentioned in Sec.~\ref{sec:appr3-module2}.}
    \label{fig:enter-label}
\end{figure}
\paragraph{Training Details: }Same data split used for training $\mathcal{E}$ and $\mathcal{D}$ is used for training $\mathcal{D}_\psi$ as well. We use Adam optimizer with initial learning rate of $1e-3$ and run the training for $60$ epochs. 
\section{Additional Results}
\subsection{Pre-trained Transition and Style Embeddings}
\begin{figure}
    \centering
    \includegraphics[width=\linewidth]{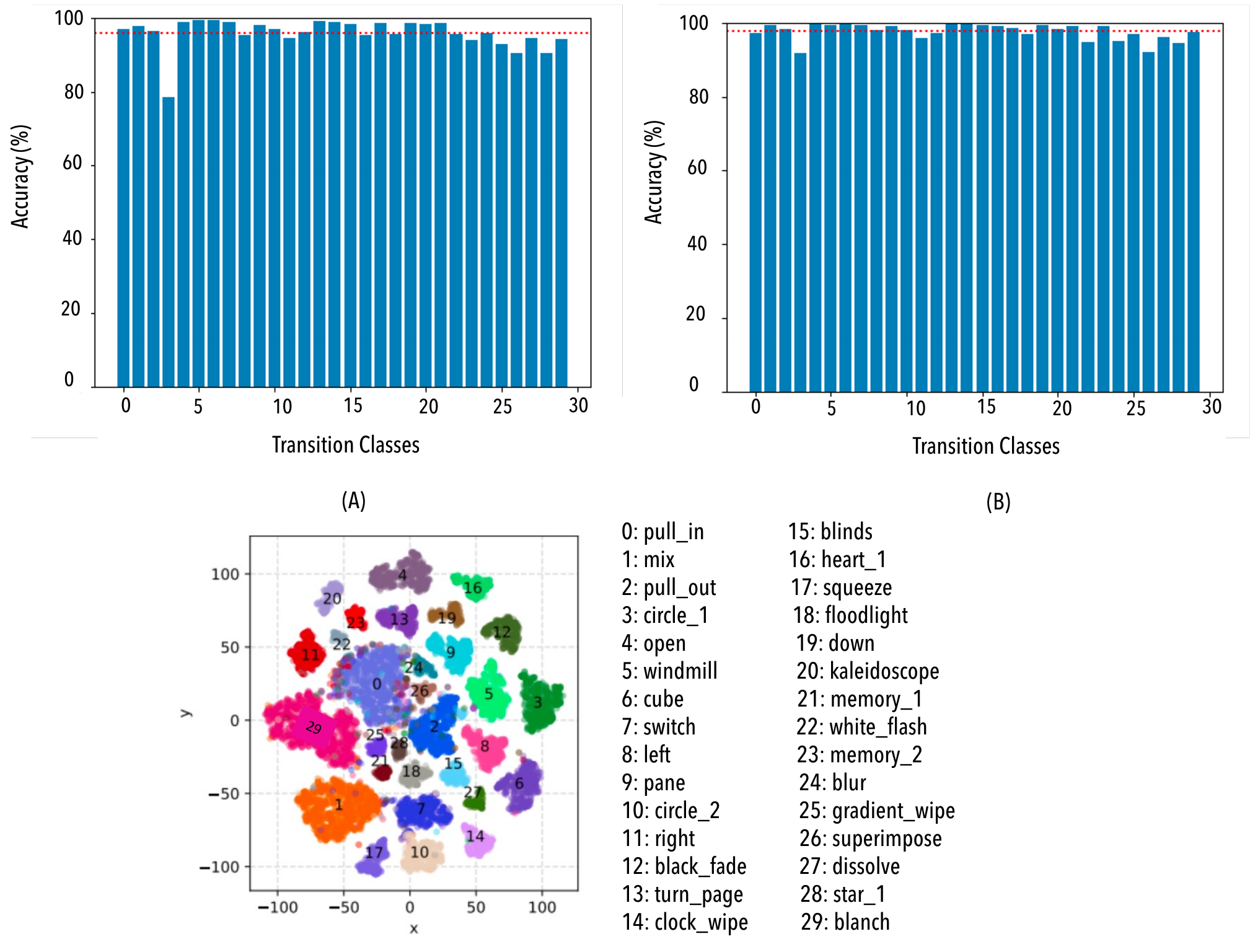}
    \caption{Transition class-wise testing accuracy obtained (A) using $\mathcal{MLN}$ (B) using AutoTransition~\cite{shen2022autotransition}'s transition classifier. As observed, adding the style conditioning doesn't affect the class-wise accuracies much.}
    \label{fig:transition_class_wise_acc}
\end{figure}

\begin{figure}
    \centering
    \includegraphics[width=0.5\linewidth]{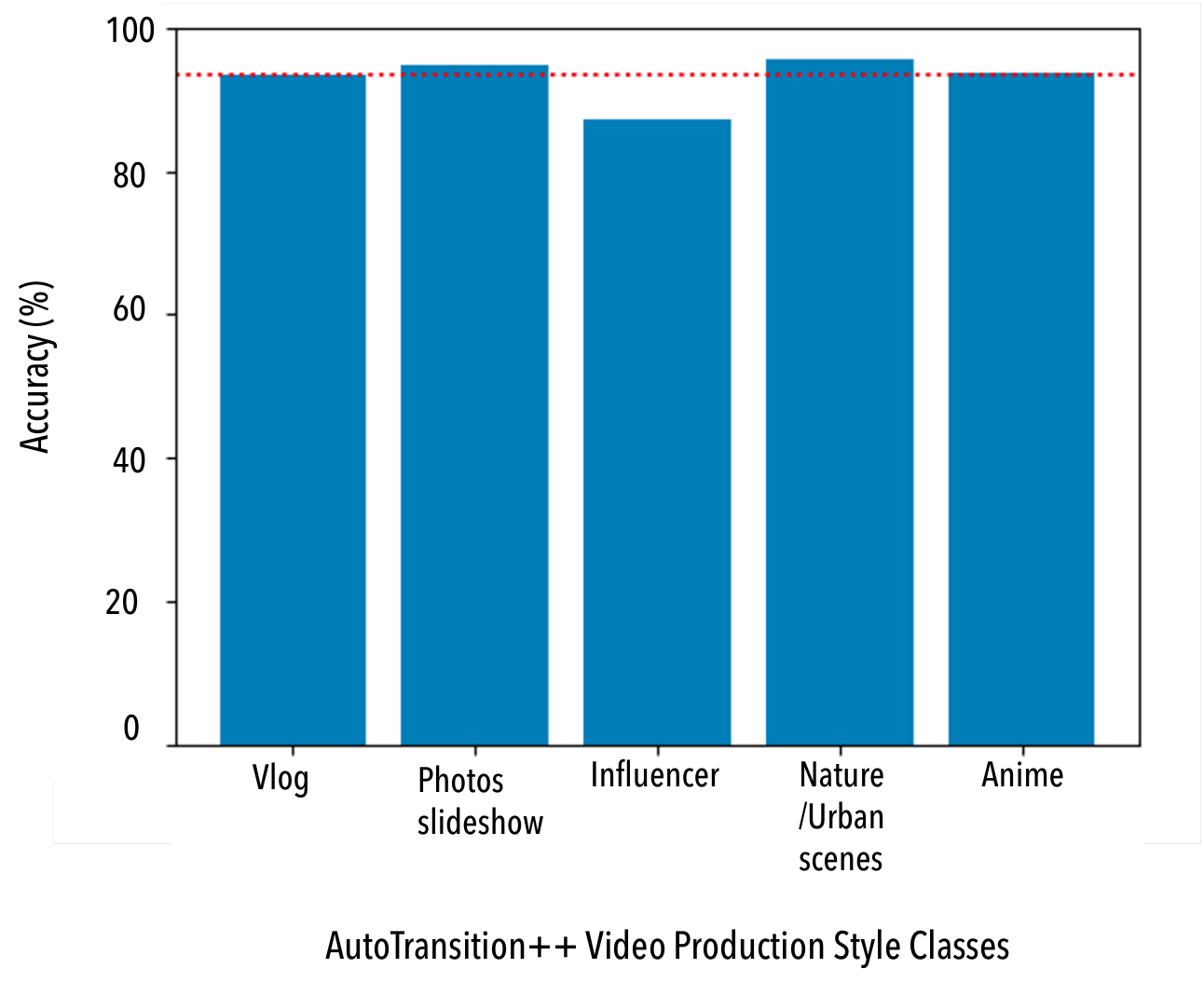}
    \caption{Video production style-wise testing accuracy obtained using $\mathcal{MLN}$ }
    \label{fig:style_class_wise_acc}
\end{figure}

Fig~\ref{fig:transition_class_wise_acc}~and Fig.~\ref{fig:style_class_wise_acc}~ show the class-wise accuracies obtained for transition classification and video style classification respectively using the multitask model $\mathcal{MLN}$. 
\subsection{Quantitative Results}
\subsubsection{Testing on more videos.}
Table~\ref{res:SCM} in Sec.~\ref{sec:expr_encode_decode} showed results for randomly selected $100$ input videos. In Table~\ref{res:SCM_full} (Appendix), we perform the same experiment on all videos that other components of~\modelname~have been tested on. We compute the cosine similarity of the mean of the embeddings obtained corresponding to the transitions recommended by the different methods and the desired production style embedding. The similarity score is computed for each video in the test dataset. The numbers reported in the table is the mean of the cosine similarity scores obtained for every video across each style. We take the mean of the transition embeddings as we are interested in understanding how good the entire sequence of transitions is with respect to the desired production style. 

\begin{table*}[t]
    \centering
    \caption{\textbf{Quantitative Results - SCM.} We show the quantitative results (cosine similarity) of our algorithm to recommend transitions to enable the adaptation of video production styles with baselines. Bold denotes \textbf{best} while underline denotes \underline{second-best}. The higher the similarity better it is.} 
    \label{tab:full_data_similarity}
    \resizebox{\textwidth}{!}{%
    \begin{tabular}{clccccc}
    \toprule
    \multirow{2}{*}{\textbf{Method}} & \multicolumn{5}{c}{\textbf{Video Production Styles}} \\
    \cmidrule{3-7}
    & & Vlog & Anime & Influencer & Photos & Nature \\
    \midrule
    \multirow{1}{*}{AutoTransition~\cite{shen2022autotransition}} & & -0.0034 &0.0009 & 0.0018 & 0.0097 & -0.0031 \\
    \multirow{1}{*}{~\modelname ($\mathcal{E}$ \& $\mathcal{D}$) }& & -0.0003 & -0.0337 & -0.0219 & -0.0076 & 0.0037 \\
    \multirow{1}{*}{~\modelname ($\mathcal{E}$ \& $\mathcal{D}$, $\mathcal{SCM}$ w/o RRT) }& & \underline{0.0194} & \underline{0.0047} & \underline{0.02} & \underline{0.0196} & \underline{0.0286} \\
    \multirow{1}{*}{~\modelname ($\mathcal{E}$ \& $\mathcal{D}$, $\mathcal{SCM}$ w/ RRT) }& & \textbf{0.2345} & \textbf{0.0117} & \textbf{0.127} & \textbf{0.728} & \textbf{0.353} \\
    \bottomrule
    \end{tabular}
    }
    \label{res:SCM_full}
\end{table*}
\subsubsection{User Studies}
\begin{figure}
    \centering
    \includegraphics[width=\columnwidth]{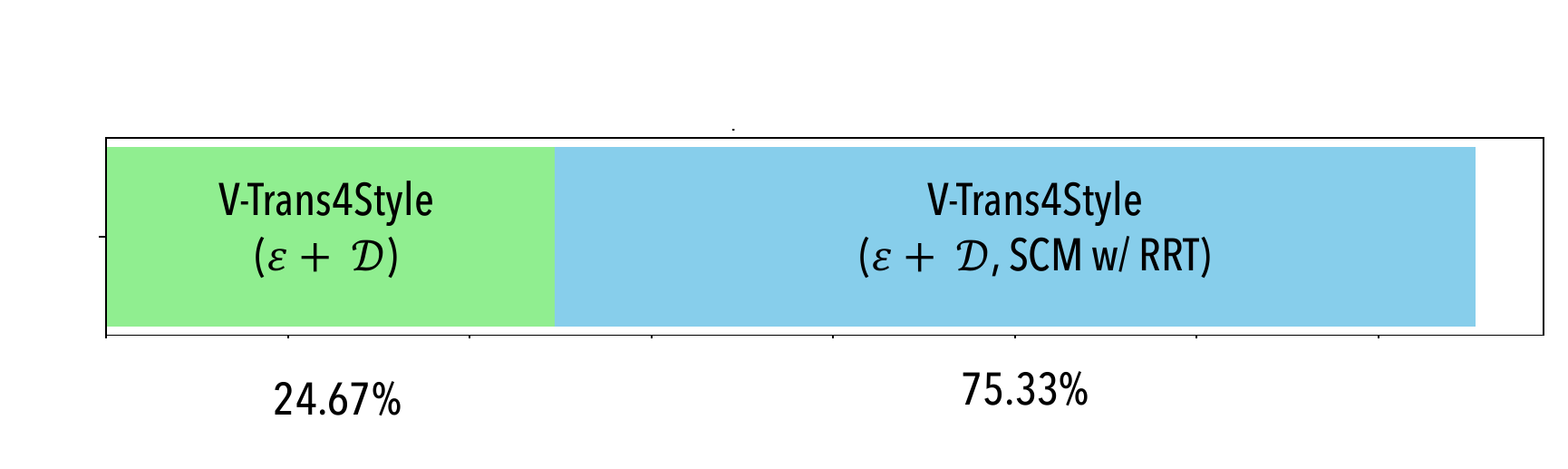}
    \caption{User Study Results.}
    \label{fig:user_study}
\end{figure}

To verify the credibility of our results further, we conducted a user study. The goal was to determine if the input videos show signs of adaptation to a different video production style after using our method's recommended transitions In this study, we engaged $102$ users with five sets of examples, each checking the adaptation to a different video production style. Each set comprised of two reference videos showcasing the desired production style and two video options. One option resulted after using visual transitions recommended by~\modelname($\mathcal{E}+\mathcal{D}, \mathcal{SCM}$ w/ RRT)~whereas the other was created using visual transitions recommended by~\modelname($\mathcal{E}+\mathcal{D}$). Both the video options were developed using the same sets of video clips. The reference videos however were different. 
The results depicted in Fig.~\ref{fig:user_study} highlight our method's capability to imbue desired video production style traits into videos originally of a different production style. All user studies have been conducted after IRB review. 
\subsection{Qualitative Analysis}
\begin{figure*}
    \centering
    \includegraphics[width=\textwidth]{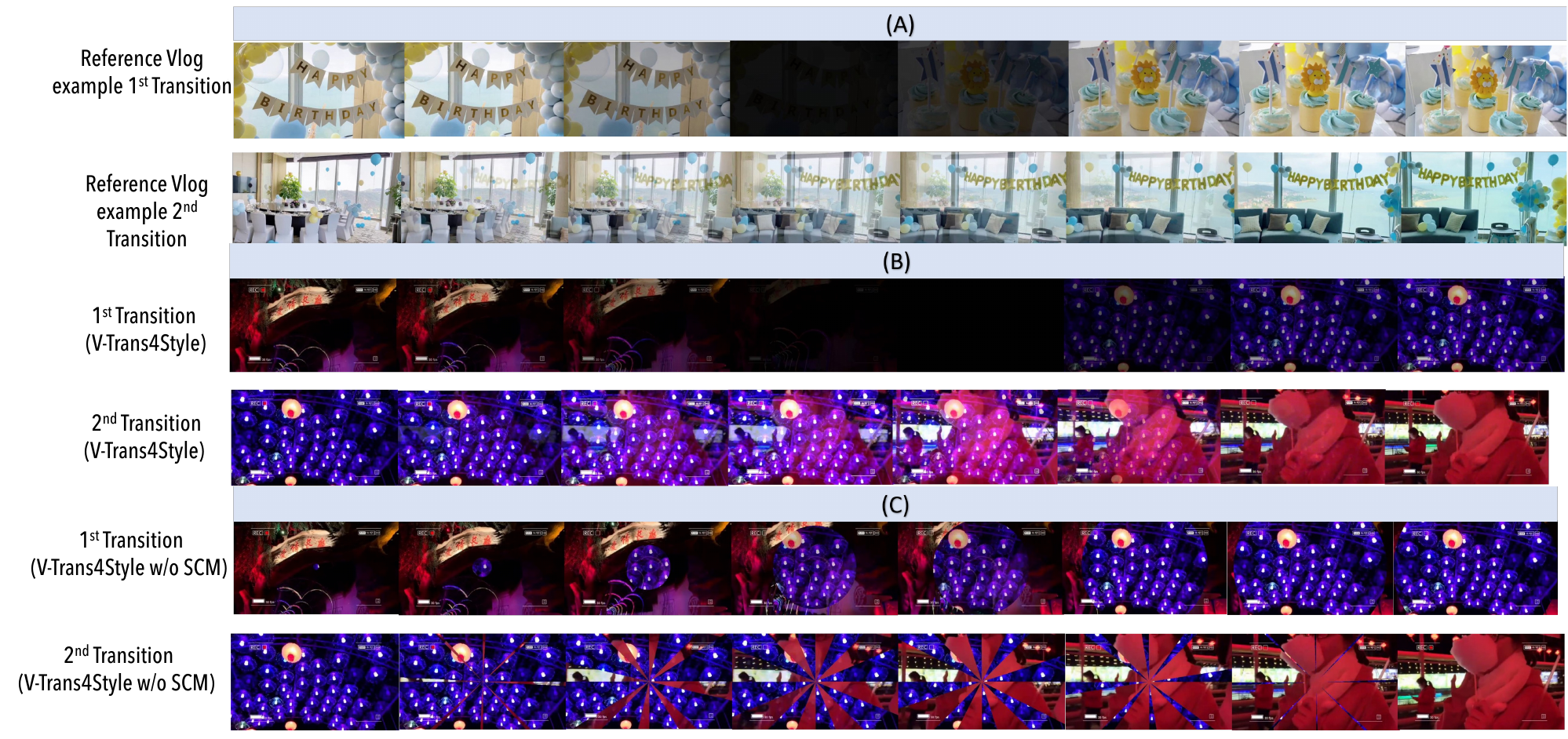}
    \caption{Set (A) corresponding to a reference video showcasing the ``vlog" video style. This is provided only to visually show the transitions that can occur in this kind of video style. The input to our model is however only the class label. Set (B) corresponds to the video obtained after applying the transitions recommended by~\modelname. Set (C) corresponds to the video obtained after applying the transitions recommended simply by~\modelname($\mathcal{E}+\mathcal{D}$). }
    \label{fig:vlogexample}
\end{figure*}

\begin{figure*}
    \centering
    \includegraphics[width=\textwidth]{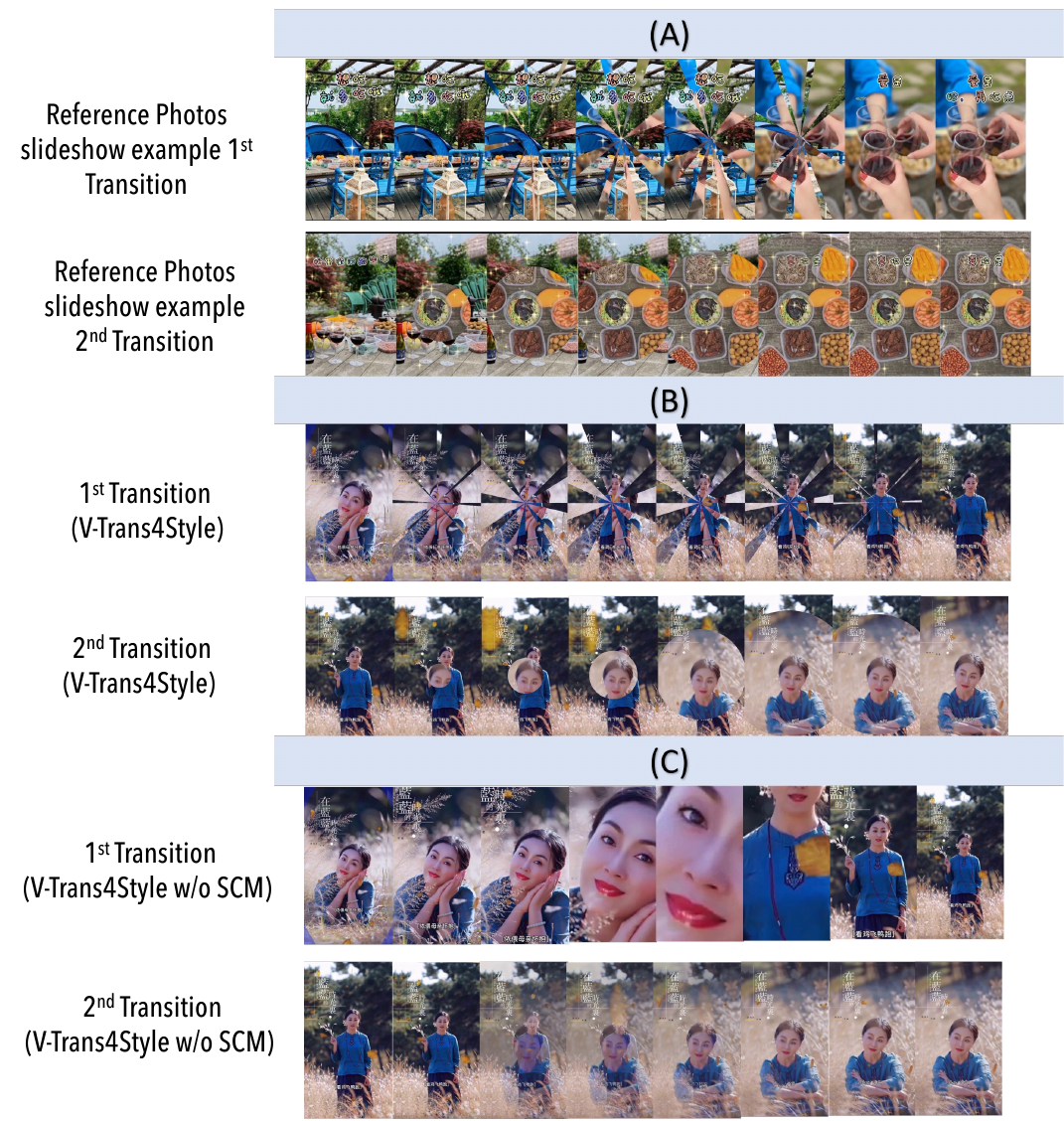}
    \caption{Set (A) corresponding to a reference video showcasing the ``photos slideshow" video style. This is provided only to visually show the transitions that can occur in this kind of video style. The input to our model is however only the class label. Set (B) corresponds to the video obtained after applying the transitions recommended by~\modelname. Set (C) corresponds to the video obtained after applying the transitions recommended simply by~\modelname($\mathcal{E}+\mathcal{D}$).}
    \label{fig:photos_slideshowexample}
\end{figure*}

\begin{figure*}
    \centering
    \includegraphics[width=\linewidth]{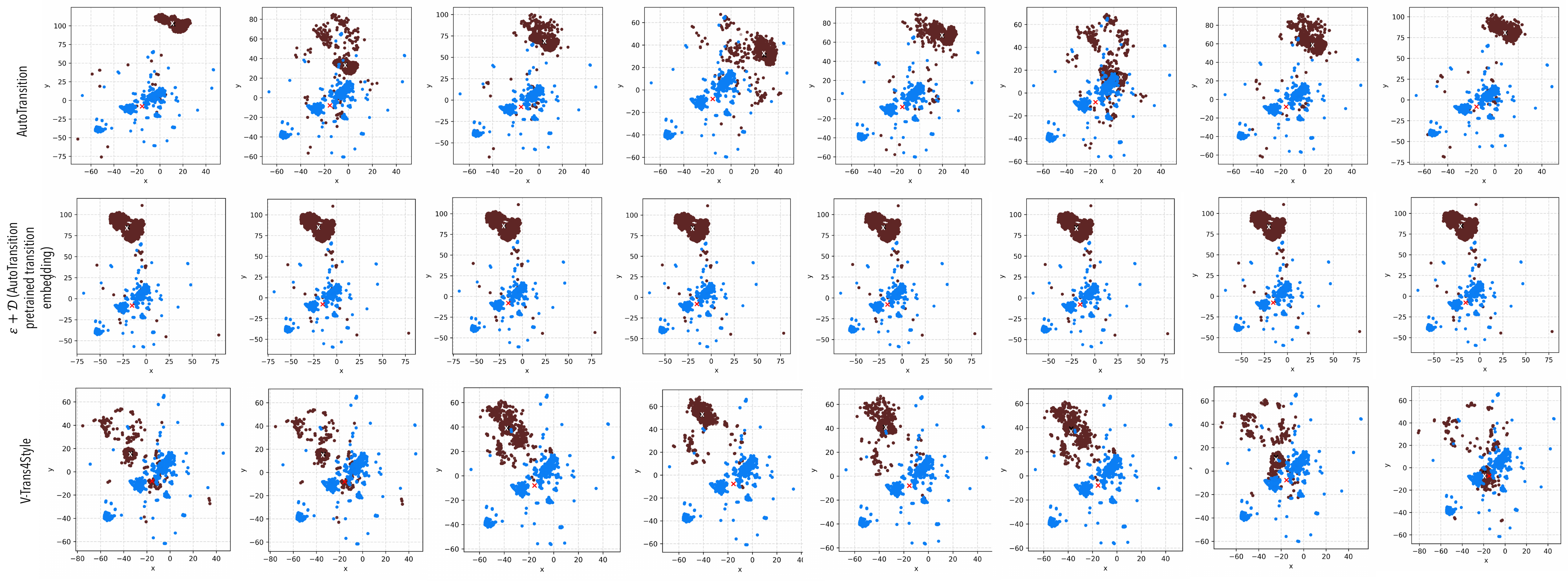}
    \caption{We took a video and adapted it to the Vlog production style. The figure here shows the changes in transition embedding space with every transition recommended in the sequence. We do this comparison across $3$ methods - AutoTransition, ~\modelname ($\mathcal{E}+\mathcal{D}$) trained on AutoTransition pre-trained transition embeddings and our~\modelname. Here, blue corresponds to the Vlog style embedding space while Brown corresponds to the mean transition embedding space.We can observe that in case of~\modelname, the mean transition embedding progressively moves towards the style embedding.Note that the progression of the transitions happen from left to right. Red cross mark corresponds to the centroid of the style embedding space. White cross mark denotes the centroid of the mean transition embedding space.}
    \label{fig:transition_traj}
\end{figure*}

We also show some qualitative results obtained from our model. We show the first few transitions in each case as the videos are long and contain my clips and visual transitions in case of Fig~\ref{fig:vlogexample} and~\ref{fig:photos_slideshowexample}. Fig~\ref{fig:transition_traj} shows the changes occurring in transitions being selected in a sequential manner to eventually achieve the desired video production style. We can see that the centroid of the transition embedding space (black cross mark) in case of~\modelname~moves closer to the style embedding space centroid (red cross mark).
\section{Broad Impact}
Video production style adaptation addresses the growing demand for personalized and engaging content. In today's digital age, with video content everywhere, viewers seek content that matches their preferences. The traditional one-size-fits-all approach often falls short, failing to cater to diverse tastes. Adapting production styles allows content creators to better serve varied audiences, including novice creators, making content creation more inclusive. This approach fosters creativity and innovation by experimenting with different styles. Our work introduces \modelname, a method that recommends visual transitions to adapt videos to different production styles. While complete adaptation involves changing elements like camera motions, cuts, colors, and audio, our work lays a strong foundation for future research. Future work will explore interactions between different editing elements and develop algorithms to streamline the adaptation process. However, it's important to consider potential negative impacts, such as content homogenization and privacy concerns. Balancing benefits and risks is crucial to ensure ethical content creation, promote diversity and inclusion, and prioritize user well-being in the digital media ecosystem.

